\documentclass[11pt,a4paper]{article}
\usepackage[hyperref]{acl2018}
\usepackage{times}
\usepackage{latexsym}
\usepackage{verbatim}
\usepackage{amssymb}
\usepackage{amsmath}
\usepackage{booktabs}
\usepackage{multicol}
\usepackage{multirow}
\usepackage{graphicx}
\usepackage{url}
\graphicspath{figures}
\usepackage{varwidth}
\usepackage{floatrow}
\usepackage{caption}
\usepackage{subcaption}

\usepackage{tikz}
\def\checkmark{\tikz\fill[scale=0.4](0,.35) -- (.25,0) -- (1,.7) -- (.25,.15) -- cycle;} 
\makeatletter
\newcommand\footnoteref[1]{\protected@xdef\@thefnmark{\ref{#1}}\@footnotemark}
\makeatother

\definecolor{orange}{RGB}{255,100,0}
\definecolor{red}{RGB}{255,0,0}
\definecolor{blue}{RGB}{0,0,255}
\definecolor{green}{RGB}{0,155,0}

\aclfinalcopy 
\def\aclpaperid{938} 

\newcommand{\topk}{{\tt Top k}}
\newcommand{\dyn}{{\tt Dyn}}
\newcommand{\topone}{{\tt Top 1}}
\newcommand{\toptwo}{{\tt Top 2}}
\newcommand{\topfour}{{\tt Top 4}}

\newcommand{\full}{\textsc{Full}}
\newcommand{\ours}{\textsc{Minimal}}
\newcommand{\oracle}{\textsc{Oracle}}

\newcommand{\fullshort}{(\textsc{Full})}
\newcommand{\oursshort}{(\textsc{Minimal})}
\newcommand{\oracleshort}{(\textsc{Oracle})}

\newcommand{\fulllong}{the standard QA model given the {\em full} document}
\newcommand{\ourslong}{the QA model given the {\em minimal} set of sentences}

\newcommand{\tfidf}{TF-IDF}
\newcommand{\ourselector}{our selector}
\newcommand{\ourselectorcap}{Our selector}

\newcommand{\venndiagram}{Venn diagram}

\newcommand{\red}[1]{\textcolor{red}{#1}}
\newcommand{\green}[1]{\textcolor{green}{#1}}
\newcommand{\blue}[1]{\textcolor{blue}{#1}}

\newcommand{\reduce}{\vspace*{-3pt}}

\title{
Efficient and Robust Question Answering \\
from Minimal Context over Documents 
}

\author{
	Sewon Min$^{1}$\thanks{All work was done while the author was an intern at Salesforce Research.}, Victor Zhong$^{2}$, Richard Socher$^{2}$, Caiming Xiong$^{2}$ \\
    Seoul National University$^{1}$, Salesforce Research$^{2}$ \\
    {\tt shmsw25@snu.ac.kr}, {\tt \{vzhong, rsocher, cxiong\}@salesforce.com}
}
\date{}

\begin{document}
\maketitle


\begin{abstract}
Neural models for question answering (QA) over documents have achieved significant performance improvements. Although effective, these models do not scale to large corpora due to their complex modeling of interactions between the document and the question. Moreover, recent work has shown that such models are sensitive to adversarial inputs. In this paper, we study the minimal context required to answer the question, and find that most questions in existing datasets can be answered with a small set of sentences. Inspired by this observation, we propose a simple sentence selector to select the minimal set of sentences to feed into the QA model. Our overall system achieves significant reductions in training (up to 15 times) and inference times (up to 13 times), with accuracy comparable to or better than the state-of-the-art on SQuAD, NewsQA, TriviaQA and SQuAD-Open. Furthermore, our experimental results and analyses show that our approach is more robust to adversarial inputs.
\end{abstract}

\section{Introduction}\label{sec:intro}\begin{table*}[ht]
\begin{center}
\resizebox{\columnwidth}{!}{
\begin{tabular}{|l|l|l|l|l|} 
 \hline
 N & $\%$ on & $\%$ on & \multirow{2}{*}{Document} & \multirow{2}{*}{Question} \\
sent & SQuAD & TriviaQA & & \\
 \hline
 1 & 90 & 56 	& {\bf In 1873, Tesla returned to his birthtown, \red{Smiljan}.} Shortly after he arrived, (...) & Where did Tesla return to in 1873? \\
  \hline
 2 & 6 & 28 		& After leaving Edison's company Tesla partnered with two businessmen in 1886, & What did Tesla Electric Light \& Manufacturing\\
 &&			&Robert Lane and Benjamin Vail, who agreed to finance an electric lighting & do? \\
 &&			& company in Tesla's name, Tesla Electric Light \& Manufacturing. {\bf The company} & \\
 &&			& {\bf \red{installed electrical arc light based illumination systems} designed by Tesla and}& \\
 &&			& {\bf also had designs for dynamo electric machine commutators, (...)} & \\
  \hline
 3$\uparrow$ & 2 & 4 		& {\bf \red{Kenneth Swezey}, a journalist whom Tesla had befriended, confirmed that Tesla} & Who did Tesla call in the middle of the night?\\
 &&			&{\bf rarely slept .} Swezey recalled one morning when Tesla called him at 3 a.m. : "I & \\
 &&			& was sleeping in my room (...) Suddenly, the telephone ring awakened me ... &\\
  \hline
 N/A & 2  & 12	& {\bf Writers whose papers are in the library are as diverse as \red{Charles Dickens} and} & The papers of which famous English Victorian \\
 &&			& {\bf Beatrix Potter.} Illuminated manuscripts in the library dating from (...) & author are collected in the library?\\
 \hline
\end{tabular}
}
\end{center}
\reduce
\caption{ Human analysis of the context required to answer questions on SQuAD and TriviaQA. 50 examples from each dataset are sampled randomly.
`N sent' indicates the number of sentences required to answer the question, and `N/A' indicates the question is not answerable even given all sentences in the document.
`Document' and `Question' are from the representative example from each category on SQuAD. Examples on TriviaQA are shown in Appendix~\ref{sec:app-analysis}.
The groundtruth answer span is in \red{red text}, and the oracle sentence (the sentence containing the grountruth answer span) is in {\bf bold text}.
} 
\label{tab:squad-motivation}
\end{table*}

\begin{table*}[ht]
\begin{center}
\resizebox{\columnwidth}{!}{
\begin{tabular}{|l|l|l|l|l|} 
 \hline
 No. & Description & \% & Sentence & Question \\
 \hline
 0 & Correct (Not exactly same & 58 & {\bf {\underline {Gothic}} architecture} is represented in the majestic churches but also at the burgher &What type of architecture is represented\\
   &as grountruth)&&houses and fortifications. &in the majestic churches?  \\
  \hline
 1 & Fail to select precise span & 6 & Brownlee argues that disobedience in opposition to the decisions of non-governmental & Brownlee argues disobedience can be \\
  &&&agencies such as {\underline {{\bf trade unions, banks, and private} universities}} can be justified if it& justified toward what institutions? \\
    &&& reflects `a larger challenge to the legal system that permits those decisions to be taken;. & \\
  \hline
 2 & Complex semantics in  & 34 &Newton was limited by Denver's defense, which sacked him {\bf seven} times and forced him& How many times did the Denver defense \\
  &sentence/question&& into {\underline {three}} turnovers, including a fumble which they recovered for a touchdown.  & force Newton into turnovers?\\
  \hline
 3 & Not answerable even with & 2 &He encourages a distinction between lawful protest demonstration, {\bf nonviolent} civil & What type of civil disobedience is \\
   &full paragraph &&disobedience, and {\underline {violent}} civil disobedience. & accompanied by aggression? \\
 \hline
\end{tabular}
}
\end{center}
\reduce
\caption{ Error cases (on exact match (EM)) of DCN+ given oracle sentence on SQuAD. 50 examples are sampled randomly. Grountruth span is in {\underline {underlined text}}, and model's prediction is in {\bf bold text}.} 
\label{tab:oracle-error}
\end{table*}


The task of textual question answering (QA), in which a machine reads a document and answers a question, is an important and challenging problem in natural language processing.
Recent progress in performance of QA models has been largely due to the variety of available QA datasets~\cite{mctest,cnndailymail,squad,newsqa,triviaqa,narrativeqa}.

Many neural QA models have been proposed for these datasets, the most successful of which tend to leverage coattention or bidirectional attention mechanisms that build codependent representations of the document and the question~\cite{dcn+,bidaf}.

Yet, learning the full context over the document is challenging and inefficient.
In particular, when the model is given a long document, or multiple documents, learning the full context is intractably slow and hence difficult to scale to large corpora.
In addition, \citet{squad-adversarial} show that, given adversarial inputs, such models tend to focus on wrong parts of the context and produce incorrect answers.

In this paper, we aim to develop a QA system that is scalable to large documents as well as robust to adversarial inputs.
First, we study the context required to answer the question by sampling examples in the dataset and carefully analyzing them.
We find that most questions can be answered using a few sentences, without the consideration of context over entire document.
In particular, we observe that on the SQuAD dataset~\cite{squad}, $92\%$ of answerable questions can be answered using a single sentence.

Second, inspired by this observation, we propose a sentence selector to select the minimal set of sentences to give to the QA model in order to answer the question.
Since the minimum number of sentences depends on the question, our sentence selector chooses a different number of sentences for each question, in contrast with previous models that select a fixed number of sentences.
Our sentence selector leverages three simple techniques --- weight transfer, data modification and score normalization, which we show to be highly effective on the task of sentence selection.

We compare \fulllong~\fullshort~and \ourslong~\oursshort~on five different QA tasks with varying sizes of documents.
On SQuAD, NewsQA, TriviaQA(Wikipedia) and SQuAD-Open, \ours~achieves significant reductions in training and inference times (up to $15\times$ and $13\times$, respectively), with accuracy comparable to or better than \full. On three of those datasets, this improvements leads to the new state-of-the-art.
In addition, our experimental results and analyses show that our approach is more robust to adversarial inputs. On the development set of SQuAD-Adversarial~\cite{squad-adversarial}, \ours~outperforms the previous state-of-the-art model by up to $13\%$.

\section{Task analyses}\label{sec:motivation}Existing QA models focus on learning the context over different parts in the full document.
Although effective, learning the context within the full document is challenging and inefficient.
Consequently, we study the minimal context in the document required to answer the question.

\subsection{Human studies}

First, we randomly sample $50$ examples from the SQuAD development set, and analyze the minimum number of sentences required to answer the question, as shown in Table~\ref{tab:squad-motivation}.
We observed that $98\%$ of questions are answerable given the document. The remaining 2\% of questions are not answerable even given the entire document.
For instance, in the last example in Table~\ref{tab:squad-motivation}, the question requires the background knowledge that Charles Dickens is an English Victorian author.
Among the answerable examples, $92\%$ are answerable with a single sentence, $6\%$ with two sentences, and $2\%$ with three or more sentences.

We perform a similar analysis on the TriviaQA (Wikipedia) development (verified) set.
Finding the sentences to answer the question on TriviaQA is more challenging than on SQuAD, since TriviaQA documents are much longer than SQuAD documents ($488$ vs $5$ sentences per document).
Nevertheless, we find that most examples are answerable with one or two sentences --- among the $88\%$ of examples that are answerable given the full document, $95\%$ can be answered with one or two sentences.

\subsection{Analyses on existing QA model}

\begin{figure}[!tb]
\centering
\resizebox{\columnwidth}{!}{
\includegraphics[width=\textwidth]{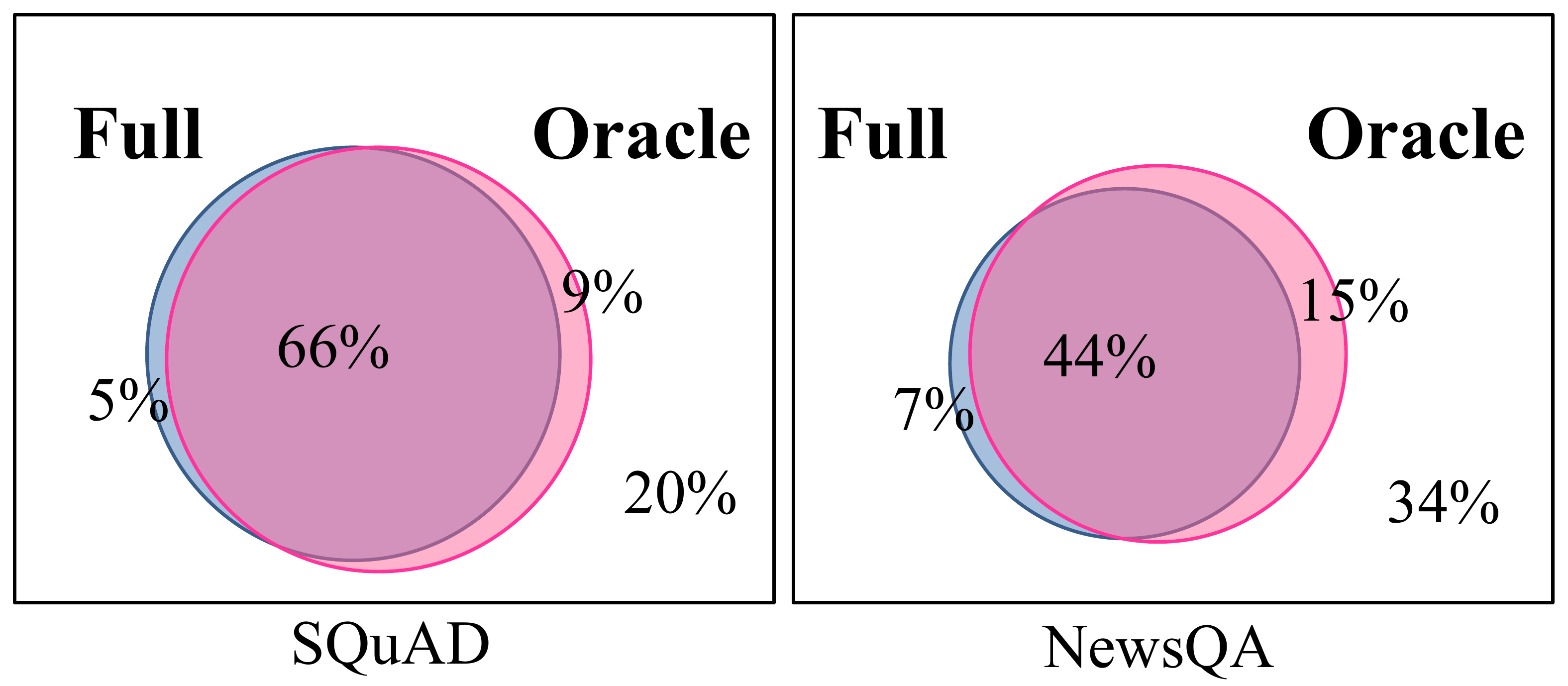}
}
\caption{
\venndiagram~of the questions answered correctly (on exact match (EM)) by the model given a full document \fullshort~and the model given an oracle sentence \oracleshort~on SQuAD (left) and NewsQA (right).
}
\label{fig:full-and-oracle}
\end{figure}

Given that the majority of examples are answerable with a single oracle sentence on SQuAD, we analyze the performance of an existing, competitive QA model when it is given the oracle sentence.
We train DCN+~\citep{dcn+}, one of the state-of-the-art models on SQuAD (details in Section~\ref{sec:method-qa-model}), on the oracle sentence.
The model achieves $83.1$ F1 when trained and evaluated using the full document and $85.1$ F1 when trained and evaluated using the oracle sentence.
We analyze 50 randomly sampled examples in which the model fails on exact match (EM) despite using the oracle sentence.
We classify these errors into 4 categories, as shown in Table~\ref{tab:oracle-error}.
In these examples, we observed that $40\%$ of questions are answerable given the oracle sentence but the model unexpectedly fails to find the answer.
$58\%$ are those in which the model's prediction is correct but does not lexically match the groundtruth answer, as shown in the first example in Table~\ref{tab:oracle-error}.
$2\%$ are those in which the question is not answerable even given the full document.
In addition, we compare predictions by the model trained using the full document \fullshort~with the model trained on the oracle sentence \oracleshort.
Figure~\ref{fig:full-and-oracle} shows the \venndiagram~of the questions answered correctly by \full~and \oracle~on SQuAD and NewsQA. \oracle~is able to answer $93\%$ and $86\%$ of the questions correctly answered by \full~on SQuAD and NewsQA, respectively.

These experiments and analyses indicate that if the model can accurately predict the oracle sentence, the model should be able to achieve comparable performance on overall QA task. 
Therefore, we aim to create an effective, efficient and robust QA system which only requires a single or a few sentences to answer the question.

\section{Method}\label{sec:method}\begin{figure*}[!tb]
\centering
\resizebox{\columnwidth}{!}{
\includegraphics[width=0.95\textwidth]{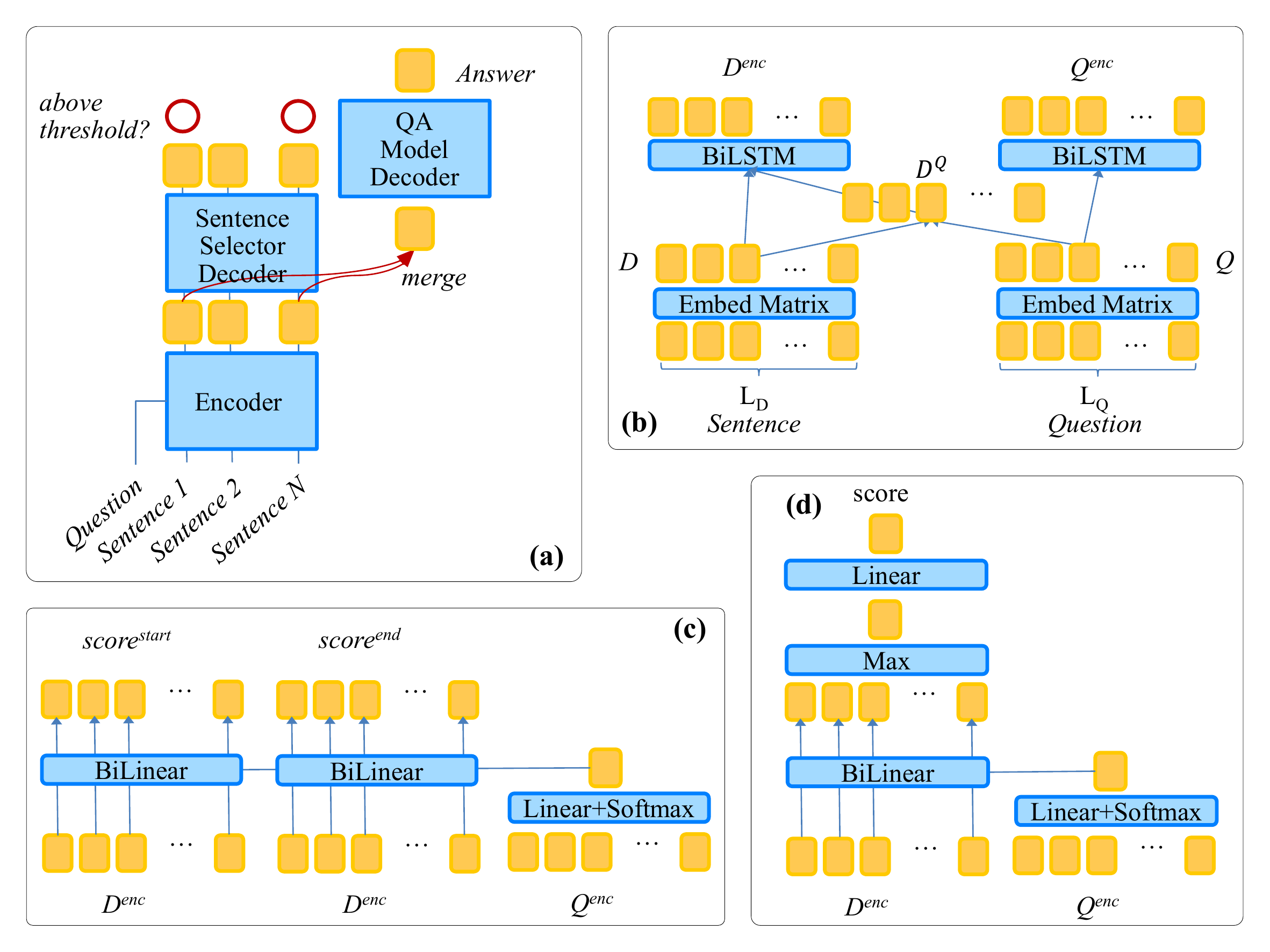}
}
\caption{
Our model architecture.
(a) Overall pipeline, consisting of sentence selector and QA model. Selection score of each sentence is obtained in parallel, then sentences with selection score above the threshold are merged and fed into QA model.
(b) Shared encoder of sentence selector and S-Reader (QA Model), which takes document and the question as inputs and outputs the document encodings $D^{enc}$ and question encodings $Q^{enc}$.
(c) Decoder of S-Reader (QA Model), which takes $D^{enc}$ and $Q^{enc}$ as inputs and outputs the scores for start and end positions.
(d) Decoder of sentence selector, which takes $D^{enc}$ and $Q^{enc}$ for each sentence and outputs the score indicating if the question is answerable given the sentence.
}
\label{fig:model-diagram}
\end{figure*}

Our overall architecture (Figure~\ref{fig:model-diagram}) consists of a sentence selector and a QA model. The sentence selector computes a selection score for each sentence in parallel. We give to the QA model a reduced set of sentences with high selection scores to answer the question.

\subsection{Neural Question Answering Model}\label{sec:method-qa-model}
We study two neural QA models that obtain close to state-of-the-art performance on SQuAD.
{\bf DCN+}~\cite{dcn+} is one of the start-of-the-art QA models, achieving $83.1$ F1 on the SQuAD development set. It features a deep residual coattention encoder, a dynamic pointing decoder, and a mixed objective that combines cross entropy loss with self-critical policy learning.
{\bf S-Reader} is another competitive QA model that is simpler and faster than DCN+, with $79.9$ F1 on the SQuAD development set.
It is a simplified version of the reader in DrQA~\cite{squad-open}, which obtains $78.8$ F1 on the SQuAD development set.
Model details and training procedures are shown in Appendix~\ref{sec:app-details}.

\subsection{Sentence Selector}\label{sec:method-sent-sel}
Our sentence selector scores each sentence with respect to the question in parallel. The score indicates whether the question is answerable with this sentence.

The model architecture is divided into the encoder module and the decoder module. The encoder is a shared module with S-Reader, which computes sentence encodings and question encodings from the sentence and the question as inputs.
First, the encoder computes sentence embeddings $D \in \mathbb{R}^{h_d \times L_d}$, question embeddings $Q \in \mathbb{R}^{h_d \times L_q}$, and question-aware sentence embeddings $D^q \in \mathbb{R}^{h_d \times L_d}$, where $h_d$ is the dimension of word embeddings, and $L_d$ and $L_q$ are the sequence length of the document and the question, respectively. Specifically, question-aware sentence embeddings are obtained as follows.

\vspace{-.5cm}
\begin{eqnarray}\label{eq:encoder-1}
    \alpha_{i} &=& \mathrm{softmax}({D_i^T}{W_1}{Q}) \in \mathbb{R}^{L_q}\\
    D^q_i &=& \sum_{j=1}^{L_q} (\alpha_{i,j}Q_j) \in \mathbb{R}^{h_d}
\end{eqnarray}

Here, $D_i\in \mathbb{R}^{h_d}$ is the hidden state of sentence embedding for the $i_{th}$ word and ${W_1} \in \mathbb{R}^{h_d \times h_d}$ is a trainable weight matrix.
After this, sentence encodings and question encodings are obtained using an LSTM~\cite{lstm}.

\vspace{-.5cm}
\begin{eqnarray}\label{eq:encoder-2}
    D^{enc} &=& \mathrm{BiLSTM}([D_i;D^q_i]) \in \mathbb{R}^{h \times L_d} \\
    Q^{enc} &=& \mathrm{BiLSTM}(Q_j) \in \mathbb{R}^{h \times L_q}
\end{eqnarray}

Here, `$;$' denotes the concatenation of two vectors, and $h$ is a hyperparameter of the hidden dimension.

Next, the decoder is a task-specific module which computes the score for the sentence by calculating bilinear similarities between sentence encodings and question encodings as follows.

\vspace{-.5cm}
\begin{eqnarray}
	\beta &=& \mathrm{softmax} (w^T Q^{enc}) \in \mathbb{R}^{L_q} \\
	 {\tilde {q^{enc}}} &=& \sum_{j=1}^{L_q} (\beta_{j}Q^{enc}_j) \in \mathbb{R}^{h}
\end{eqnarray}

\vspace{-2cm}
\begin{eqnarray}
     {\tilde {h_i}} &=& (D^{enc}_i{W_2}{\tilde {q^{enc}}}) \in \mathbb{R}^{h} \\
     {\tilde {h}} &=& \mathrm{max}({\tilde {h_1}}, {\tilde {h_2}}, \cdots, {\tilde {h_{L_d}}})\\
    score &=& {W_3^T}{\tilde {h}} \in \mathbb{R}^2
\end{eqnarray}

Here, $w \in \mathbb{R}^h, W_2 \in \mathbb{R}^{h \times h \times h}, W_3 \in \mathbb{R}^{h \times 2}, $ are trainable weight matrices. Each dimension in $score$ means the question is answerable or nonanswerable given the sentence.



\begin{table*}[ht]
\begin{center}
\resizebox{0.8\columnwidth}{!}{
\begin{tabular}{|l|l|l|l|l|l|} 
 \hline
 Dataset & Domain & N word & N sent & N doc & Supervision \\
 \hline
 SQuAD & Wikipedia & 155 & 5 & - & Span \\
 NewsQA & News Articles & 803 & 20 & - & Span \\
 TriviaQA (Wikipedia) & Wikipedia & 11202 & 488 & 2 & Distant \\
 SQuAD-Open & Wikipedia & 120734 & 4488 & 10 & Distant \\
 SQuAD-Adversarial-AddSent & Wikipedia & 169 & 6 & - & Span \\
 SQuAD-Adversarial-AddOneSent & Wikipedia & 165 & 6 & - & Span \\
 \hline
\end{tabular}
}
\end{center}
\reduce
\caption{ Dataset used for experiments. `N word', `N sent' and `N doc' refer to the average number of words, sentences and documents, respectively.
All statistics are calculated on the development set. 
For SQuAD-Open, since the task is in open-domain, we calculated the statistics based on top 10 documents from Document Retriever in DrQA~\citep{squad-open}.} 
\label{tab:dataset}
\vspace{-8pt}
\end{table*}

We introduce 3 techniques to train the model.
(i) As the encoder module of our model is identical to that of S-Reader, we transfer the weights to the encoder module from the QA model trained on the single oracle sentence \oracleshort.
(ii) We modify the training data by treating a sentence as a wrong sentence if the QA model gets $0$ F1, even if the sentence is the oracle sentence.
(iii) After we obtain the score for each sentence, we normalize scores across sentences from the same paragraph, similar to \citet{simple-and-effective}. 
All of these three techniques give substantial improvements in sentence selection accuracy, as shown in Table~\ref{tab:class-result}.
More details including hyperparameters and training procedures are shown in Appendix~\ref{sec:app-details}.

Because the minimal set of sentences required to answer the question depends on the question, we select the set of sentences by thresholding the sentence scores, where the threshold is a hyperparameter (details in Appendix~\ref{sec:app-details}).
This method allows the model to 
select a variable number of sentences for each question, as opposed to a fixed number of sentences for all questions. Also, by controlling the threshold, the number of sentences can be dynamically controlled during the inference. We  define \dyn~(for Dynamic) as this method, and define \topk~as the method which simply selects the top-$k$ sentences for each question.

\section{Experiments}\label{sec:exp}\subsection{Dataset and Evaluation Metrics}\label{sec:exp-dataset}
We train and evaluate our model on five different datasets as shown in Table~\ref{tab:dataset}.\\

\vspace{-.2cm}
\paragraph{SQuAD}~\cite{squad} is a well-studied QA dataset on Wikipedia articles that requires each question to be answered from a paragraph.

\vspace{-.2cm}
\paragraph{NewsQA}~\cite{newsqa} is a dataset on news articles that also provides a paragraph for each question, but the paragraphs are longer than those in SQuAD.

\vspace{-.2cm}
\paragraph{TriviaQA}~\cite{triviaqa} is a dataset on a large set of documents from the Wikipedia domain and Web domain. Here, we only use the Wikipedia domain.
Each question is given a much longer context in the form of multiple documents.

\vspace{-.2cm}
\paragraph{SQuAD-Open}~\cite{squad-open} is an open-domain question answering dataset based on SQuAD. In SQuAD-Open, only the question and the answer are given. The model is responsible for identifying the relevant context from all English Wikipedia articles.

\vspace{-.2cm}
\paragraph{SQuAD-Adversarial}~\cite{squad-adversarial} is a variant of SQuAD. It shares the same training set as SQuAD, but an adversarial sentence is added to each paragraph in a subset of the development set. \\

We use accuracy (Acc) and mean average precision (MAP) to evaluate sentence selection.
We also measure the average number of selected sentences (N sent) to compare the efficiency of our \dyn~method and the \topk~method. 

To evaluate the performance in the task of question answering, we measure F1 and EM (Exact Match), both  being standard metrics for evaluating span-based QA.
In addition, we measure training speed (Train Sp) and inference speed (Infer Sp) relative to the speed of standard QA model \fullshort.
The speed is measured using a single GPU (Tesla K80), and includes the training and inference time for the sentence selector.


\begin{table}[ht]
\begin{center}
\resizebox{\columnwidth}{!}{
\begin{tabular}{|l||c|c||c|c|c|} 
 \hline
 \multirow{2}{*}{Model}& \multicolumn{2}{c||}{SQuAD} & \multicolumn{3}{c|}{NewsQA} \\
\cline{2-6}
& Top 1 & MAP & Top 1 & Top 3 & MAP \\
 \hline
 \tfidf & 81.2 & 89.0 		& 49.8 & 72.1 & 63.7 \\
 \ourselectorcap & 85.8 & 91.6 		& 63.2 & 85.1 & 75.5 \\
 \ourselectorcap~(T) & 90.0 & 94.3  	& 67.1 & 87.9 & 78.5 \\
 \ourselectorcap~(T+M, T+M+N) & {\bf 91.2} & {\bf 95.0} 	& {\bf 70.9} & {\bf 89.7} & {\bf 81.1} \\
 \hline
\citet{squad-selection-sota} & - & 92.1 		& - & - & - \\
 \hline
 \end{tabular}
} 
\resizebox{0.9\columnwidth}{!}{
\begin{tabular}{|l||c|c||c|c|} 
 \hline
  \multirow{2}{*}{Selection method} & \multicolumn{2}{c||}{SQuAD} & \multicolumn{2}{c|}{NewsQA} \\
\cline{2-5}
& N sent & Acc & N sent & Acc \\
 \hline
 \topk~\small{(T+M)\footnote{\label{t}`N' does not change the result on \topk, since \topk~depends on the relative scores across the sentences from same paragraph.}}& 1 & 91.2 & 1 & 70.9 \\
 \topk~\small{(T+M)\footnoteref{t}}& 2 & 97.2 & 3 & 89.7 \\
 \topk~\small{(T+M)\footnoteref{t}}& 3 & 98.9 & 4 & 92.5 \\
  \hline
 \dyn~\small{(T+M)}& 1.5 & 94.7 & 2.9 & 84.9 \\
 \dyn~\small{(T+M)}& 1.9 & 96.5 & 3.9 & 89.4 \\
 \hline
 \dyn~\small{(T+M+N)}& 1.5 & 98.3 & 2.9 & 91.8 \\
 \dyn~\small{(T+M+N)}& 1.9 & {\bf 99.3} & 3.9 & {\bf 94.6} \\
 \hline
\end{tabular}
}
\end{center}
\caption{ Results of sentence selection on the dev set of SQuAD and NewsQA.
(Top)
We compare different models and training methods.
We report Top 1 accuracy (Top 1) and Mean Average Precision (MAP).
Our selector outperforms the previous state-of-the-art~\cite{squad-selection-sota}.
(Bottom)
We compare different selection methods. We report the number of selected sentences (N sent) and the accuracy of sentence selection (Acc).
`T', `M' and `N' are training techniques described in Section~\ref{sec:method-sent-sel} (weight transfer, data modification and score normalization, respectively).
} 
\label{tab:class-result}
\vspace{-8pt}
\end{table}

\begin{table}[ht]
\begin{center}
\resizebox{\columnwidth}{!}{
\begin{tabular}{|l||c|c|c|c|} 
 \hline
 \multicolumn{5}{|c|}{SQuAD (with S-Reader)}\\
  \hline
  & F1 & EM & Train Sp & Infer Sp \\
  \hline
 \full & 79.9 & 71.0 & x1.0 & x1.0 	\\
 \oracle & 84.3 & 74.9 & x6.7 & x5.1 	\\
 \ours(\topk) & 78.7 & 69.9 & {\bf x6.7} & {\bf x5.1}  \\
 \ours(\dyn) & 79.8 & 70.9 & {\bf x6.7} & x3.6 	\\
 \hline
 \hline
 \multicolumn{5}{|c|}{SQuAD (with DCN+)}\\
  \hline
 \full & 83.1 & 74.5 & x1.0 & x1.0 	 \\
 \oracle &  85.1 & 76.0 & x3.0 & x5.1 				\\
 \ours(\topk) & 79.2 & 70.7 & x3.0 & {\bf x5.1}			 \\
 \ours(\dyn) & 80.6 & 72.0 & x3.0 & x3.7 			\\
  \hline
 GNR & 75.0\footnote{\label{t}Numbers on the test set.} & 66.6\footnoteref{t} & - & - \\
 {FastQA} & 78.5 & 70.3 & - & -\\
 {FusionNet} & {\bf 83.6} & {\bf 75.3} & - & - \\
  \hline
  \hline
  \multicolumn{5}{|c|}{NewsQA (with S-Reader)}\\
  \hline
& F1 & EM & Train Sp & Infer Sp  \\
  \hline
 \full& 63.8 & 50.7 & x1.0 & x1.0 \\
 \oracle & 75.5 & 59.2 & x18.8 & x21.7 \\
 \ours(\topk) & 62.3 & 49.3 & {\bf x15.0} & {\bf x6.9} \\
 \ours(\dyn) & {\bf 63.2} & {\bf 50.1} & {\bf x15.0} & x5.3 \\
 \hline
  {FastQA} & 56.1 & 43.7 & - & -\\
  \hline
\end{tabular}
}
\end{center}
\reduce
\caption{ Results on the dev set of SQuAD (First two) and NewsQA (Last).
For \topk, we use $k=1$ and $k=3$ for SQuAD and NewsQA, respectively. We compare with GNR~\cite{gnr}, FusionNet~\cite{fusionnet} and FastQA~\cite{fastqa}, which are the model leveraging sentence selection for question answering, and the published state-of-the-art models on SQuAD and NewsQA, respectively.} 
\label{tab:qa-result}
\end{table}

\begin{table*}[!ht]
\begin{center}
\resizebox{\columnwidth}{!}{
\begin{tabular}{|l|} 
\hline
\vspace{-5pt} The initial LM model weighed approximately 33,3000 pounds, and allowed surface stays up to around \blue{34 hours}. \\
\vspace{-1pt}. . . \\
An Extended Lunar Module weighed over 36,200 pounds, and allowed surface stays of over \underline{\red{3 days}}. \red{\checkmark}\\
\vspace{-.2cm} \\
\emph{For about how long would the extended LM allow a surface stay on the moon?}\\
 \hline
 \hline
\vspace{-5pt} Approximately \underline{\red{1,000}} British soldiers were killed or injured. \red{\checkmark}\\
\vspace{-1pt}. . . \\
The remaining \blue{500} British troops, led by George Washington, retreated to Virginia. \\
\vspace{-.2cm} \\
\emph{How many casualties did British get?} \\
 \hline
This book, which influenced the thought of Charles Darwin, successfully promoted the doctrine of \underline{\blue{uniformitarianism}}. \\
This theory states that slow geological processes have occurred throughout the Earth's history and are still occurring today. \red{\checkmark}\\
 In contrast, \red{catastrophism} is the theory that Earth's features formed in single, catastrophic events and remained unchanged thereafter. \red{\checkmark}\\
 \vspace{-.2cm} \\
 \emph{Which theory states that slow geological processes are still occuring today, and have occurred throughout Earth's history?}\\
 \hline
\end{tabular}
}
\end{center}
\reduce
\caption{ Examples on SQuAD. Grountruth span ({\underline {underlined text}}), the prediction from \full~(\blue{blue text}) and \ours~(\red{red text}). Sentences selected by \ourselector~is denoted with \red{\checkmark}.
In the above two examples, \ours~correctly answer the question by selecting the oracle sentence. In the last example, \ours~fails to answer the question, since the inference over first and second sentences is required to answer the question.} 
\label{tab:squad-analysis}
\end{table*}

\begin{table*}[!ht]
\begin{center}
\resizebox{\columnwidth}{!}{
\begin{tabular}{|c|l|} 
\hline
 selected & sentence \\
 \hline
 \blue{\checkmark} \green{\checkmark} \red{\checkmark} & However, in 1883-84 Germany began to build a colonial empire in Africa and the South Pacific, before losing interest in imperialism. \\
\green{\checkmark} \red{\checkmark}& The establishment of the German colonial empire proceeded smoothly, starting with German New Guinea in {\underline {1884}}.
\\
 \hline
 \multicolumn{2}{|l|}{\emph{When did Germany found their first settlement?} \blue{1883-84} \green{1884} \red{1884}}\\
 \hline
 \hline
 \blue{\checkmark} \green{\checkmark} \red{\checkmark} & In the {\underline {late 1920s}}, Tesla also befriended George Sylvester Viereck, a poet, writer, mystic, and later, a Nazi propagandist. \\
 \green{\checkmark} & In middle age, Tesla became a close friend of Mark Twain; they spent a lot of time together in his lab and elsewhere. \\
 \hline
 \multicolumn{2}{|l|}{\emph{When did Tesla become friends with Viereck?} \blue{late 1920s} \green{middle age} \red{late 1920s}}\\
 \hline
\end{tabular}
}
\end{center}
\reduce
\caption{An example on SQuAD, where the sentences are ordered by the score from \ourselector. Grountruth span ({\underline {underlined text}}), the predictions from \topone~(\blue{blue text}), \toptwo~(\green{green text}) and \dyn~(\red{red text}). Sentences selected by \topone, \toptwo~and \dyn~are denoted with \blue{\checkmark}, \green{\checkmark} and \red{\checkmark}, respectively.
} 
\label{tab:dyn-over-topk}
\end{table*}

\subsection{SQuAD and NewsQA}\label{sec:squad-and-newsqa}
For each QA model, we experiment with three types of inputs. First, we use the full document \fullshort. Next, we give the model the oracle sentence containing the groundtruth answer span \oracleshort. Finally, we select sentences using our sentence selector \oursshort, using both \topk~and \dyn. 
We also compare this last method with \tfidf~method for sentence selection, which selects sentences using n-gram \tfidf~distance between each sentence and the question.

\begin{figure}[!ht]
\centering
\resizebox{\columnwidth}{!}{
\includegraphics[width=\textwidth]{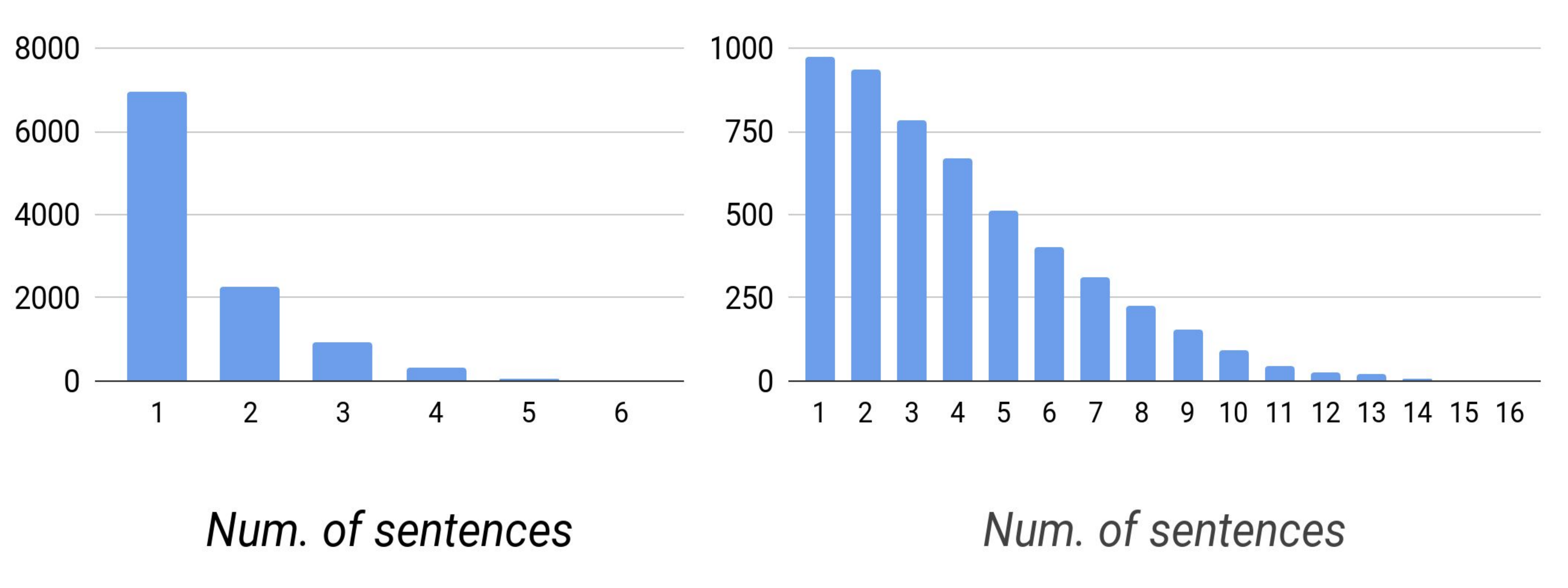}
}
\caption{
The distributions of number of sentences that \ourselector~selects using \dyn~method on the dev set of SQuAD (left) and NewsQA (right).
}
\label{fig:num-of-sentences}
\end{figure}

\paragraph{Results}
Table~\ref{tab:class-result} shows results in the task of sentence selection on SQuAD and NewsQA.
First, \ourselector~outperforms \tfidf~method and the previous state-of-the-art by large margin (up to $2.9\%$ MAP).

Second, our three training techniques -- weight transfer, data modification and score normalization -- improve performance by up to $5.6\%$ MAP. 
Finally, our \dyn~method achieves higher accuracy with less sentences than the \topk~method.
For example, on SQuAD, \toptwo~achieves $97.2$ accuracy, whereas \dyn~achieves $99.3$ accuracy with 1.9 sentences per example.
On NewsQA, \topfour~achieves $92.5$ accuracy, whereas \dyn~achieves $94.6$ accuracy with 3.9 sentences per example. 

Figure~\ref{fig:num-of-sentences} shows that the number of sentences selected by \dyn~method vary substantially on both SQuAD and NewsQA.
This shows that \dyn~chooses a different number of sentences depending on the question, which reflects our intuition.

Table~\ref{tab:qa-result} shows results in the task of QA on SQuAD and NewsQA.
\ours~is more efficient in training and inference than \full.
On SQuAD, S-Reader achieves $6.7\times$ training and $3.6\times$ inference speedup on SQuAD, and $15.0\times$ training and $6.9\times$ inference speedup on NewsQA.
In addition to the speedup, \ours~achieves comparable result to \full~(using S-Reader, $79.9$ vs $79.8$ F1 on SQuAD and $63.8$ vs $63.2$ F1 on NewsQA).

We compare the predictions from \full~and \ours~in Table~\ref{tab:squad-analysis}.
In the first two examples, our sentence selector chooses the oracle sentence, and the QA model correctly answers the question.
In the last example, our sentence selector fails to choose the oracle sentence, so the QA model cannot predict the correct answer. 
In this case, \ourselector~chooses the second and the third sentences instead of the oracle sentence because the former contains more information relevant to question. In fact, the context over the first and the second sentences is required to correctly answer the question.

Table~\ref{tab:dyn-over-topk} shows an example on SQuAD, which \ours~with \dyn~correctly answers the question, and \ours~with \topk~sometimes does not. \topone~selects one sentence in the first example, thus fails to choose the oracle sentence. \toptwo~selects two sentences in the second example, which is inefficient as well as leads to the wrong answer. In both examples, \dyn~selects the oracle sentence with minimum number of sentences, and subsequently predicts the answer. More analyses are shown in Appendix~\ref{sec:app-analysis}.

\subsection{TriviaQA and SQuAD-Open}\label{sec:triviaqa-and-squad-open}
TriviaQA and SQuAD-Open are QA tasks that reason over multiple documents.
They do not provide the answer span and only provide the question-answer pairs.

\begin{table*}[!ht]
\begin{center}
\resizebox{0.85\columnwidth}{!}{
\begin{tabular}{|c|c||c|c|c|c|c||c|c|c|c|c|}
 \hline
\multicolumn{2}{|c||}{\multirow{2}{*}{}} & \multicolumn{5}{c||}{TriviaQA (Wikipedia)}  & \multicolumn{5}{c|}{SQuAD-Open}  \\
\cline{3-12}
\multicolumn{2}{|c||}{} & n sent & Acc & Sp & F1 & EM  & n sent & Acc & Sp & F1 & EM \\
  \hline
 \multicolumn{2}{|c||}{\full} & 69 & 95.9 & x1.0 & 59.6 & 53.5 	& 124 & 76.9 & x1.0 & 41.0 & 33.1 \\
 \hline
 \multirow{4}{*}{\ours} &  \multirow{2}{*}{\tfidf} 
 			& 5 & 73.0 & x13.8 & 51.9 & 45.8  	&5&46.1&x12.4	& 36.6 & 29.6\\
 			&& 10 & 79.9 & x6.9 & 57.2 & 51.5 	&10&54.3&x6.2	& 39.8 & 32.5\\
 \cline{2-12}
 & Our 	& 5.0 & 84.9 & {\bf x13.8} & 59.5 & 54.0 &5.3&58.9&{\bf x11.7}	& {\bf 42.3}&{\bf 34.6} \\
 & Selector & 10.5 & 90.9 & x6.6  & {\bf 60.5} & {\bf 54.9} &10.7&64.0&x5.8 & {\bf 42.5}&{\bf 34.7} \\
 \hline
 \multicolumn{2}{|c||}{Rank 1} &-&-&-  & 56.0\footnoteref{t} & 51.6\footnoteref{t} &2376\footnote{\label{a}Approximated based on there are 475.2 sentences per document, and they use 5 documents per question}&77.8&-&-& 29.8 \\
 \multicolumn{2}{|c||}{Rank 2} &-&-&-& 55.1\footnoteref{t} & 48.6\footnoteref{t}  &-&-&-& 37.5 & 29.1\\
 \multicolumn{2}{|c||}{Rank 3} &-&-&-& 52.9\footnote{\label{t}Numbers on the test set.} & 46.9\footnoteref{t} &2376\footnoteref{a}&77.8&-&-& 28.4 \\
 \hline
\end{tabular}
}
\end{center}
\caption{Results on the dev-full set of TriviaQA (Wikipedia) and the dev set of SQuAD-Open.
Full results (including the dev-verified set on TriviaQA) are in Appendix~\ref{sec:app-results}.
For training \full~and \ours~on TriviaQA, we use $10$ paragraphs and $20$ sentences, respectively.
For training \full~and \ours~on SQuAD-Open, we use $20$ paragraphs and $20$ sentences, respectively.
For evaluating \full~and \ours, we use $40$ paragraphs and $5$-$20$ sentences, respectively.
`n sent' indicates the number of sentences used during inference.
`Acc' indicates accuracy of whether answer text is contained in selected context.
`Sp' indicates inference speed.
We compare with the results from the sentences selected by \tfidf~method and our selector (\dyn).
We also compare with published Rank1-3 models. For TriviaQA(Wikipedia), they are Neural Casecades~\cite{neural-cascades}, Reading Twice for Natural Language Understanding~\cite{readingtwice} and Mnemonic Reader~\cite{mnemonic}.
For SQuAD-Open, they are DrQA~\cite{squad-open} (Multitask), R$^3$~\cite{rthree} and DrQA (Plain).
} 
\label{tab:large-qa-result}
\end{table*}

For each QA model, we experiment with two types of inputs.
First, since TriviaQA and SQuAD-Open have many documents for each question, we first filter paragraphs based on the \tfidf~similarities between the question and the paragraph, and then feed the full paragraphs to the QA model \fullshort.
On TriviaQA, we choose the top 10 paragraphs for training and inference.
On SQuAD-Open, we choose the top 20 paragraphs for training and the top 40 for inferences.
Next, we use our sentence selector with \dyn~\oursshort.
We select $5$-$20$ sentences using our sentence selector, from $200$ sentences based on \tfidf.

For training the sentence selector, we use two techniques described in Section~\ref{sec:method-sent-sel}, weight transfer and score normalization, but we do not use data modification technique, since there are too many sentences to feed each of them to the QA model.
For training the QA model, we transfer the weights from the QA model trained on SQuAD, then fine-tune.

\paragraph{Results}
Table~\ref{tab:large-qa-result} shows results on TriviaQA (Wikipedia) and SQuAD-Open.
First, \ours~obtains higher F1 and EM over \full, with the inference speedup of up to $13.8\times$.
Second, the model with our sentence selector with \dyn~achieves higher F1 and EM over the model with \tfidf~selector.
For example, on the development-full set, with $5$ sentences per question on average, the model with \dyn~achieves $59.5$ F1 while the model with \tfidf~method achieves $51.9$ F1.
Third, we outperforms the published state-of-the-art on both dataset.

\subsection{SQuAD-Adversarial}\label{sec:squad-adv}
We use the same settings as Section~\ref{sec:squad-and-newsqa}.
We use the model trained on SQuAD, which is exactly same as the model used for Table~\ref{tab:qa-result}. For \ours, we select top 1 sentence from our sentence selector to the QA model.

\begin{table}[ht]
\begin{center}
\resizebox{\columnwidth}{!}{
\begin{tabular}{|l|l||c|c|c||c|c|c|} 
 \hline
 \multicolumn{2}{|c||}{\multirow{2}{*}{SQuAD-Adversarial}} & \multicolumn{3}{c||}{AddSent} & \multicolumn{3}{c|}{AddOneSent} \\
 \cline{3-8}
 \multicolumn{2}{|c||}{} & F1 & EM & Sp & F1 & EM & Sp \\
 \hline
 \multirow{3}{*}{DCN+} & \full & 52.6 & 46.2 & x0.7 & 63.5 & 56.8 & x0.7\\
 & \oracle & 84.2 & 75.3 & x4.3 & 84.5 & 75.8 & x4.3\\
 & \ours & {\bf 59.7} & {\bf 52.2} & x4.3 & {\bf 67.5} & {\bf 60.1} & x4.3\\
  \hline
  \multirow{3}{*}{S-Reader} & \full & 57.7 & 51.1 & x1.0 & 66.5 & 59.7 & x1.0 \\
 & \oracle & 82.5 & 74.1 & x6.0 & 82.9 & 74.6 & x6.0 \\
 & \ours & 58.5 & 51.5 & {\bf x6.0} & 66.5 & 59.5 & {\bf x6.0}  \\
 \hline
  \multicolumn{2}{|c||}{RaSOR} & 39.5 &-&-& 49.5 &-&- \\
   \multicolumn{2}{|c||}{ReasoNet} & 39.4 &-&-& 50.3&-&- \\
  \multicolumn{2}{|c||}{Mnemonic Reader} & 46.6 &-&-& 56.0 &-&- \\
  \hline
\end{tabular}
}
\end{center}
\caption{ Results on the dev set of SQuAD-Adversarial. We compare with RaSOR~\cite{rasor}, ReasoNet~\cite{reasonet} and Mnemonic Reader~\cite{mnemonic}, the previous state-of-the-art on SQuAD-Adversarial, where the numbers are from \citet{squad-adversarial}.} 
\label{tab:adv-result}
\end{table}

\begin{table*}[ht]
\begin{center}
\resizebox{\columnwidth}{!}{
\begin{tabular}{|l|} 
 \hline
San Francisco mayor \underline{\red{Ed Lee}} said of the highly visible homeless presence in this area "they are going to have to leave".\\
\blue{Jeff Dean} was the mayor of Diego Diego during Champ Bowl 40.\\
 \hline
 \em{Who was the mayor of San Francisco during Super Bowl 50?}\\
  \hline
   \hline
In January 1880, two of Tesla's uncles put together enough money to help him leave Gospić for \underline{\red{Prague}} where he was to study.\\
 Tadakatsu moved to the city of \blue{Chicago} in 1881.\\
 \hline
 \em{What city did Tesla move to in 1880?}\\
 \hline
\end{tabular}
}
\end{center}
\reduce
\caption{Examples on SQuAD-Adversarial. Groundtruth span is in {\underline {underlined text}}, and predictions from \full~and \ours~are in \blue{blue text} and \red{red text}, respectively.} 
\label{tab:adv-analysis}
\vspace{-8pt}
\end{table*}

\paragraph{Results}
Table~\ref{tab:adv-result} shows that \ours~outperforms \full, achieving the new state-of-the-art by large margin ($+11.1$ and $+11.5$ F1 on AddSent and AddOneSent, respectively).

Figure~\ref{tab:adv-analysis} compares the predictions by DCN+ \full~(blue) and \ours~(red).
While \full~selects the answer from the adversarial sentence, \ours~first chooses the oracle sentence, and subsequently predicts the correct answer.
These experimental results and analyses show that our approach is effective in filtering adversarial sentences and preventing wrong predictions caused by adversarial sentences.

\section{Related Work}\label{sec:related}\paragraph{Question Answering over Documents}

There has been rapid progress in the task of question answering (QA) over documents along with various datasets and competitive approaches.
Existing datasets differ in the task type, including multi-choice QA~\cite{mctest}, cloze-form QA~\cite{cnndailymail} and extractive QA~\cite{squad}.
In addition, they cover different domains, including Wikipedia~\cite{squad,triviaqa}, news~\cite{cnndailymail,newsqa}, fictional stories~\cite{mctest,narrativeqa}, and textbooks~\cite{race,cloth}.

Many neural QA models have successfully addressed these tasks by leveraging coattention or bidirectional attention mechanisms~\cite{dcn+,bidaf} to model the codependent context over the document and the question.
However, \citet{squad-adversarial} find that many QA models are sensitive to adversarial inputs.

Recently, researchers have developed large-scale QA datasets, which requires answering the question over a large set of documents in a closed~\cite{triviaqa} or open-domain~\cite{searchqa,webquestions,squad-open,quasar}.
Many models for these datasets either retrieve documents/paragraphs relevant to the question~\cite{squad-open,simple-and-effective,rthree}, or leverage simple non-recurrent architectures to make training and inference tractable over large corpora~\cite{neural-cascades,fast-and-accurate}.

\paragraph{Sentence selection} 
The task of selecting sentences that can answer to the question has been studied across several QA datasets~\cite{wikiqa}, by modeling relevance between a sentence and the question~\cite{wikiqa1,wikiqa2,qatransfer}.
Several recent works also study joint sentence selection and question answering.
\citet{coarse2fine} propose a framework that identifies the sentences relevant to the question (property) using simple bag-of-words representation, then generates the answer from those sentences using recurrent neural networks.
\citet{gnr} cast the task of extractive question answering as a search problem by iteratively selecting the sentences, start position and end position. 
They are different from our work in that (i) we study of the minimal context required to answer the question, (ii) we choose the minimal context by selecting variable number of sentences for each question, while they use a fixed size of number as a hyperparameter, (iii) our framework is flexible in that it does not require end-to-end training and can be combined with existing QA models, and (iv) they do not show robustness to adversarial inputs.

\section{Conclusion}\label{sec:concl}We proposed an efficient and robust QA system that is scalable to large documents and robust to adversarial inputs.
First, we studied the minimal context required to answer the question in existing datasets and found that most questions can be answered using a small set of sentences.
Second, inspired by this observation, we proposed a sentence selector which selects a minimal set of sentences to answer the question to give to the QA model.
We demonstrated the efficiency and effectiveness of our method across five different datasets with varying sizes of source documents.
We achieved the training and inference speedup of up to $15\times$ and $13\times$, respectively, and accuracy comparable to or better than existing state-of-the-art.
In addition, we showed that our approach is more robust to adversarial inputs.

\section*{Acknowledgments}
We thank the anonymous reviewers and the Salesforce Research team members for their thoughtful comments and discussions.

\bibliography{00-main}
\bibliographystyle{acl_natbib}

\clearpage

\appendix
\section{Models Details}\label{sec:app-details}\begin{figure*}[pht]
\centering
\resizebox{\columnwidth}{!}{
\includegraphics[width=\textwidth]{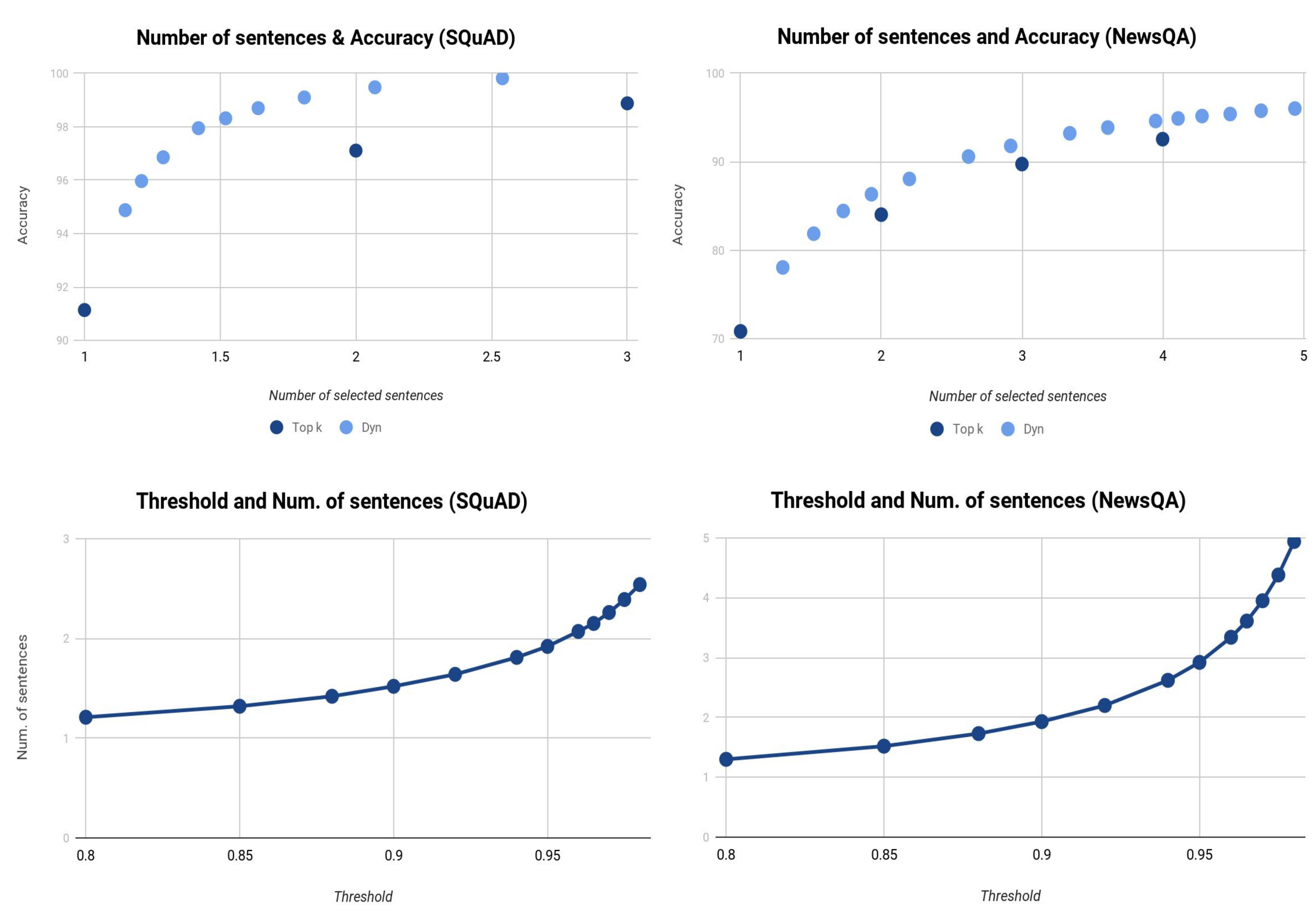}
}
\caption{
(Top) The trade-off between the number of selected sentence and accuracy on SQuAD and NewsQA. \dyn~outperforms \topk~in accuracy with similar number of sentences. (Bottom) Number of selected sentences depending on threshold.
}
\label{fig:thresholding}
\end{figure*}

\paragraph{S-Reader}
The model architecture of S-Reader is divided into the encoder module and the decoder module. The encoder module is identical to that of our sentence selector. It first takes the document and the question as inputs, obtains document embeddings $D \in \mathbb{R}^{L_d \times h_d}$, question embeddings $Q \in \mathbb{R}^{L_q \times h_d}$ and question-aware document embeddings $D^q \in \mathbb{R}^{L_d \times h_d}$, where $D^q$ is defined as Equation~\ref{eq:encoder-1}, and finally obtains document encodings $D^{enc}$ and question encodings $Q^{enc}$ as Equation~\ref{eq:encoder-2}.
The decoder module obtains the scores for start and end position of the answer span by calculating bilinear similarities between document encodings and question encodings as follows.

\vspace{-.5cm}
\begin{eqnarray}
	\beta &=& \mathrm{softmax} (w_1^T Q^{enc}) \in \mathbb{R}^{L_q} \\
	 {\tilde {q^{enc}}} &=& \sum_{j=1}^{L_q} (\beta_{j}Q^{enc}_j) \in \mathbb{R}^{h} \\
    score^{start} &=& D^{enc} W_{start} {\tilde q^{enc}} \in \mathbb{R}^{L_d} \\
    score^{end} &=& D^{enc} W_{end} {\tilde q^{enc}} \in \mathbb{R}{L_d}
\end{eqnarray}

Here, $w_{1} \in \mathbb{R}^{h}, W_{start}, W_{end} \in \mathbb{R}^{h \times h}$ are trainable weight matrices.

The overall architecture is similar to Document Reader in DrQA~\cite{squad-open}, except they are different in obtaining embeddings and use different hyperparameters.
As shown in Table~\ref{tab:qa-result}, our S-Reader obtains F1 score of $79.9$ on SQuAD development data, while Document Reader in DrQA achieves $78.8$.

\paragraph{Training details}
We implement all of our models using PyTorch. First, the corpus is tokenized using Stanford CoreNLP toolkit~\citep{corenlp}. We obtain the embeddings of the document and the question by concatenating $300$-dimensional Glove embeddings pretrained on the 840B Common Crawl corpus~\citep{glove}, $100$-dimensional character n-gram embeddings by \citet{charembedding}, and $300$-dimensional contextualized embeddings pretrained on WMT~\citep{cove}.
We do not use handcraft word features such as POS and NER tagging, which is different from Document Reader in DrQA.
Hence, the dimension of the embedding ($d_h$) is 600. We use the hidden size ($h$) of $200$.
We apply dropout with 0.2 drop rate~\cite{srivastava2014dropout} to encodings and LSTMs for regularization.
We train the models using ADAM optimizer~\citep{adam} with default hyperparameters.
When we train and evaluate the model on the dataset, the document is truncated to the maximum length of $min(2000, max(1000, L_{th}))$ words, where  $L_{th}$  is the length which covers $90\%$ of documents in the whole examples.

\paragraph{Selection details}
Here, we describe how to dynamically select sentences using \dyn~method. Given the sentences $S_{all} = \{s_1, s_2, s_3, ..., s_n\}$, ordered by scores from the sentence selector in descending order, the selected sentences $S_{selected}$ is as follows.

\vspace{-.5cm}
\begin{eqnarray}
  S_{candidate} &=& \{s_i \in S_{all} | score(s_i) >= 1-th\} \\
  S_{selected} &=&
  \begin{cases}
  	S_{candidate} & if S_{candidate} = \emptyset \\
    \{s_1\} & o.w.
  \end{cases}
\end{eqnarray}

Here, $score(s_i)$ is the score of sentence $s_i$ from the sentence selector, and $th$ is a hyperparameter between $0$ and $1$.

The number of sentences to select can be dynamically controlled during inference by adjusting $th$, so that proper number of sentences can be selected depending on the needs of accuracy and speed.
Figure~\ref{fig:thresholding} shows the trade-off between the number of sentences and accuracy, as well as the number of selected sentences depending on the threshold $th$.

\section{More Analyses}\label{sec:app-analysis}\paragraph{Human studies on TriviaQA}
We randomly sample $50$ examples from the TriviaQA (Wikipedia) development (verified) set, and analyze the minimum number of sentences to answer the question.
Despite TriviaQA having longer documents ($488$ sentences per question), most examples are answerable with one or two sentences, as shown in Table~\ref{tab:triviaqa-motivation}.
While $88\%$ of examples are answerable given the full document, $95\%$ of them can be answered with one or two sentences.

\begin{figure*}[pht]
\centering
\resizebox{\columnwidth}{!}{
\includegraphics[width=\textwidth]{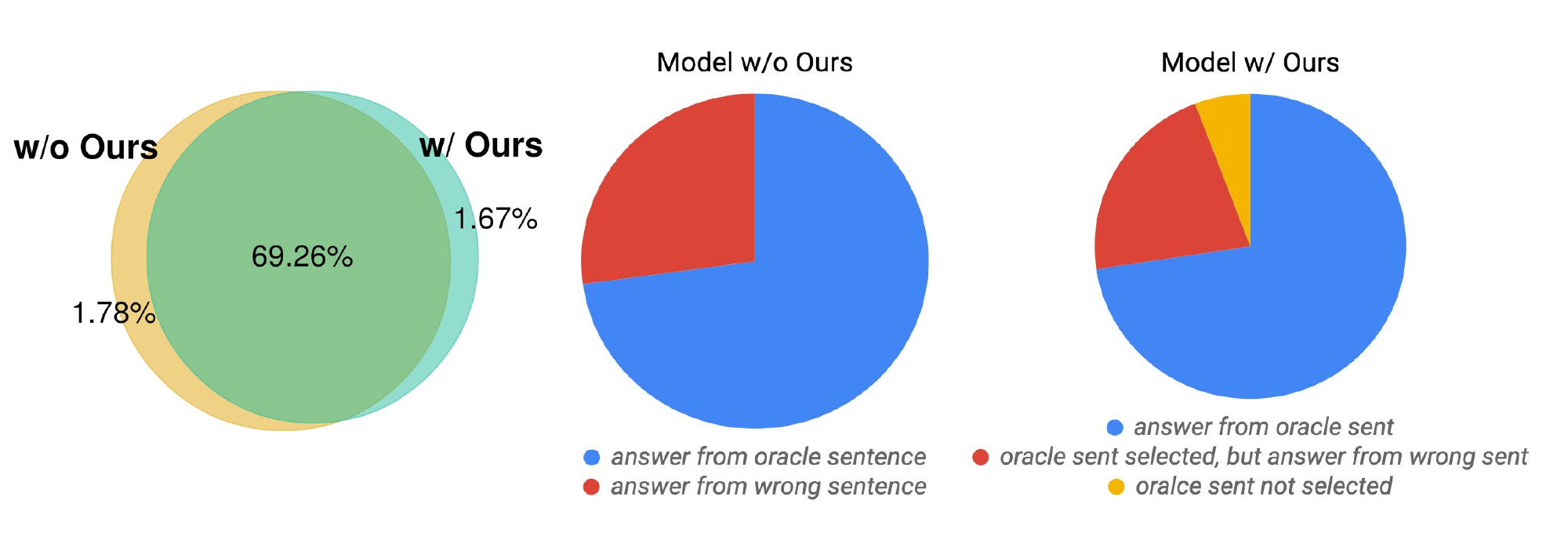}
}
\caption{
(Left) \venndiagram of the questions answered correctly by \full~and with \ours. (Middle and Right) Error cases from \full~(Middle) and \ours~(Right), broken down by which sentence the model's prediction comes from.
}
\label{fig:squad-error-analysis}
\end{figure*}

\begin{table*}[pht]
\begin{center}
\resizebox{\columnwidth}{!}{
\begin{tabular}{|l|l|l|l|} 
\hline
 N sent & $\%$ & Paragraph & Question \\
 \hline
1 & 56 & Chicago O'Hare International Airport, also known as O'Hare Airport, Chicago International Airport, Chicago & In which city would you find O'Hare\\
&&O'Hare or simply O'Hare, is an international airport located on the far northwest side of \red{Chicago}, Illinois.& International Airport?\\
\cline{3-4}
&& In 1994, Wet Wet Wet had their biggest hit, a cover version of the troggs' single "Love is All Around", which  & The song "Love is All Around" by \\
&&was used on the soundtrack to the film \red{Four Weddings and A Funeral}. &Wet Wet Wet featured on the sound-  \\
&&&track for which 1994 film?\\
 \hline
2 & 28 & Cry Freedom is a 1987 British epic drama film directed by Richard Attenborough, set in late-1970s apartheid & The 1987 film `Cry Freedom' is a \\
&& era South Africa. (...) The film centres on the real-life events involving black activist \red{Steve Biko} and (...) &biographical drama about which South \\
&&& Aftrican civil rights leader? \\
\cline{3-4}
&&Helen Adams Keller was an American author, political activist, and lecturer. (…) The story of how Keller’s teacher, & Which teacher taught Helen Keller\\
&&\red{Anne Sullivan}, broke through the isolation imposed by a near complete lack of language, allowing the girl to  & to communicate?\\
&& blossom as she learned to communicate, has become widely known through (...) 
& \\
 \hline
3 $\uparrow$ & 4 & (...) The equation shows that, as volume increases, the pressure of the gas decreases in proportion. Similarly,
 & Who gave his name to the scientific  \\
 &&as volume decreases, the pressure of the gas increases. The law was named after chemist and physicist &law that states that the pressure of a gas \\
 &&\red{Robert Boyle}, who published the original law. (...) &is inversely proportional to its \\
 &&&volume at constant temperature?\\
 \cline{3-4}
&& The \red{Buffalo six} (known primarily as Lackawanna Six
) is a group of six Yemeni-American friends who were & Mukhtar Al-Bakri, Sahim Alsan, Faysal \\
&& convicted of providing material support to Al Qaeda in December 2003, (…) In the late summer of 2002, one of &Galan, Shafal Mosed, Yaseinn Taher and  \\
&& the members, Mukhtar Al-Bakri, sent (…)  Yahya Goba and Mukhtar Al-Bakri received 10-year prison sentences. &Yahya Goba were collectively known as the\\
&& Yaseinn Taher and Shafal Mosed received 8-year prison sentences. Sahim Alwan received a 9.5-year sentence. & “Lackawanna Six” and by what other name? \\
&& Faisal Galab received a 7-year sentence. & \\
 \hline
N/A & 12 & (...) A commuter rail operation, the \red{New Mexico} Rail Runner Express, connects the state's capital, its & Which US state is nicknamed both `the\\
&& and largest city, and other communities. (...) & Colourful State' and `the Land of \\
&&&Enchantment?'\\
\cline{3-4}
&& Smith also arranged for the publication of a series of etchings of “Capricci” in his vedette ideal, &
Canaletto is famous for his landscapes \\
&& but the returns were not high enough, and in 1746 Canaletto moved to \red{London}, to be closer to his market. & of Venice and which other city? \\
 \hline
\end{tabular}
}
\end{center}
\reduce
\caption{
Human analysis of the context required to answer questions on TriviaQA (Wikipedia). 50 examples are sampled randomly. `N sent' indicates the number of sentences required to answer the question, and `N/A' indicates the question is not answerable even given all sentences in the document. The groundtruth answer text is in \red{red text}. Note that the span is not given as the groundtruth.
In the first example classified into `N/A', the question is not answerable even given whole documents, because there is no word `corlourful' or `enchantment' in the given documents. In the next example, the question is also not answerable even given whole documents, because all sentences containing `London' does not contain any information about Canaletto's landscapes.
}
\label{tab:triviaqa-motivation}
\end{table*}

\begin{table*}[!pht]
\begin{center}
\resizebox{\columnwidth}{!}{
\begin{tabular}{|l|} 
\hline
In On the Abrogation of the Private Mass, he condemned as \red{idolatry} the idea that the mass is a sacrifice, asserting instead that it is a {\underline {gift}}, to be \\ received with thanksgiving by the whole congregation. \red{\checkmark} \\
\\
\emph{What did Luther call the mass instead of sacrifice?}\\
 \hline
 \hline
Veteran receiver {\underline {Demaryius Thomas}} led the team with 105 receptions for 1,304 yards and six touchdowns, while Emmanuel Sanders caught 
(...) \red{\checkmark} \\
Running back \red{Ronnie Hillman} also made a big impact with 720 yards, five touchdowns, 24 receptions, and a 4.7 yards per carry average. \red{\checkmark} \\
\\
\emph{Who had the most receptions out of all players for the year?} \\
 \hline
In 1211, after the conquest of Western Xia, \red{Genghis Kahn} planned again to conquer the Jin dynasty. \red{\checkmark}\\
Instead, the Jin commander sent a messenger, {\underline {Ming-Tan}}, to the Mongol side, who defected and told the Mongols that the Jin army was waiting \\
on the other side of the pass. \\
The Jin dynasty collapsed in 1234, after the siege of Caizhou. \red{\checkmark}\\
\\
 \emph{Who was the Jin dynasty defector who betrayed the location of the Jin army?}\\
 \hline
\end{tabular}
}
\end{center}
\reduce
\caption{Examples on SQuAD, which \ours~predicts the wrong answer. Grountruth span is in {\underline {underlined text}}, the prediction from \ours~is in \red{red text}. Sentences selected by \ourselector~is denoted with \red{\checkmark}.
In the first example, the model predicts the wrong answer from the oracle sentence. In the second example, the model predicts the answer from the wrong sentence, although it selects the oracle sentence. In the last example, the model fails to select the oracle sentence.
} 
\label{tab:squad-analysis-more}
\end{table*}

\begin{table*}[!ht]
\begin{center}
\resizebox{0.65\columnwidth}{!}{
\begin{tabular}{|c|c||c|c|c|c|c|c|c|}
 \hline
\multicolumn{2}{|c||}{\multirow{2}{*}{TriviaQA}} & \multicolumn{3}{c|}{Inference}  & \multicolumn{2}{c|}{Dev-verified} & \multicolumn{2}{c|}{Dev-full} \\
\cline{3-9}
 \multicolumn{2}{|c||}{} & n sent & Acc & Sp & F1 & EM & F1 & EM \\
  \hline
 \multicolumn{2}{|c||}{\full} & 69 & 95.9 & x1.0 & 66.1 & 61.6 & 59.6 & 53.5\\
 \hline
 \multirow{4}{*}{\ours} &  \multirow{3}{*}{\tfidf} 
 			& 5 & 73.0 & x13.8 & 60.4 & 54.1 & 51.9 & 45.8 \\
 			&& 10 & 79.9 & x6.9 & 64.8 & 59.8 & 57.2 & 51.5 \\
 			&& 20 & 85.5 & x3.5 & 67.3 & 62.9 &60.4 & 54.8 \\
  \cline{2-9} &
  \multirow{3}{*}{Our Selector} & 5.0 & 84.9 & {\bf x13.8} &  65.0 & 61.0 & 59.5 & 54.0 \\
   && 10.5 & 90.9 & x6.6  & {\bf 67.0} & {\bf 63.8} & {\bf 60.5} & {\bf 54.9} \\
   && 20.4 & 95.3 & x3.4 & {\bf 67.7} & {\bf 63.8} & {\bf 61.3} & {\bf 55.6} \\
  \hline
  \multicolumn{2}{|c||}{MEMEN} &-&-&-& 55.8 & 49.3 & 46.9 & 43.2  \\
  \multicolumn{2}{|c||}{Mnemonic Reader} &-&-&-& 59.5\footnote{\label{t}Numbers on the test set.} & 54.5\footnoteref{t} & 52.9\footnoteref{t} & 46.9\footnoteref{t} \\
  \multicolumn{2}{|c||}{Reading Twice} &-&-&-& 59.9\footnoteref{t}& 53.4\footnoteref{t} & 55.1\footnoteref{t} & 48.6\footnoteref{t}\\
  \multicolumn{2}{|c||}{Neural Cascades} &-&-&-& 62.5\footnoteref{t} & 58.9\footnoteref{t}  & 56.0\footnoteref{t} & 51.6\footnoteref{t} \\
  \hline
\end{tabular}
}
\end{center}
\caption{
Results on the dev-verified set and the dev-full set of TriviaQA (Wikipedia).
We compare the results from the sentences selected by \tfidf~and \ourselector~(\dyn).
We also compare with MEMEN~\cite{memen}, Mnemonic Reader~\cite{mnemonic}, Reading Twice for Natural Language Understanding~\cite{readingtwice} and Neural Casecades~\cite{neural-cascades}, the published state-of-the-art.
} 
\label{tab:triviaqa-result}
\end{table*}

\begin{table*}[!ht]
\begin{center}
\resizebox{0.5\columnwidth}{!}{
\begin{tabular}{|c|c||c|c|c|c|c|c|} 
 \hline
 \multicolumn{2}{|c||}{\multirow{2}{*}{SQuAD-Open}} & \multicolumn{3}{c|}{Inference}  & \multicolumn{2}{c|}{Dev} \\
\cline{3-7}
 \multicolumn{2}{|c||}{}& n sent & Acc & Sp & F1 & EM \\
  \hline
 \multicolumn{2}{|c||}{\full} & 124 & 76.9 & x1.0 & 41.0 & 33.1 \\
 \hline
 \multirow{6}{*}{\ours} & \multirow{4}{*}{\tfidf} &5&46.1&x12.4	& 36.6 & 29.6\\
 	&&10&54.3&x6.2	& 39.8 & 32.5\\
  	&&20&62.4&x3.1 & 41.7 & 34.1\\
    &&40&65.8&x1.6 & 42.5 & 34.6\\
  \cline{2-7}
 & \multirow{2}{*}{Our} &5.3&58.9&{\bf x11.7}	& {\bf 42.3}&{\bf 34.6}\\
   	&\multirow{2}{*}{Selector}&10.7&64.0&x5.8 & {\bf 42.5}&{\bf 34.7}\\
   	&&20.4&68.1&x3.0 & {\bf 42.6}&{\bf 34.7}\\
 	&&40.0&71.4&x1.5 & 42.6&34.7\\
  \hline
  \multicolumn{2}{|c||}{R$^3$} &-&-&-& 37.5 & 29.1 \\
 \multicolumn{2}{|c||}{DrQA} &2376\footnote{\label{a}Approximated based on there are 475.2 sentences per document, and they use 5 documents per question}&77.8&-&-& 28.4 \\
 \multicolumn{2}{|c||}{DrQA (Multitask)} &2376\footnoteref{a}&77.8&-&-& 29.8 \\
  \hline
\end{tabular}
}
\end{center}
\caption{Results on the dev set of SQuAD-Open.
We compare with the results from the sentences selected by \tfidf~method and our selector (\dyn).
We also compare with R$^3$~\cite{rthree} and DrQA~\cite{squad-open}.
} 
\label{tab:squad-open-result}
\end{table*}

\begin{table*}[!ht]
\begin{center}
\resizebox{\columnwidth}{!}{
\begin{tabular}{|l|l|l|l|} 
 \hline
 Analysis & Table & Dataset & Ids \\
 \hline
 Context Analysis & \ref{tab:squad-motivation} & SQuAD & 56f7eba8a6d7ea1400e172cf, 56e0bab7231d4119001ac35c, 56dfa2c54a1a83140091ebf6, 56e11d8ecd28a01900c675f4, \\
&&& 572ff7ab04bcaa1900d76f53, 57274118dd62a815002e9a1d, 5728742cff5b5019007da247, 572748745951b619008f87b2, \\ &&& 573062662461fd1900a9cdf7, 56e1efa0e3433e140042321a,  57115f0a50c2381900b54aa9, 57286f373acd2414000df9db, \\ &&& 57300f8504bcaa1900d770d3, 57286192ff5b5019007da1e0, 571cd11add7acb1400e4c16f, 57094ca7efce8f15003a7dd7, \\ &&& 57300761947a6a140053cf9c, 571144d1a58dae1900cd6d6f,  572813b52ca10214002d9d68, 572969f51d046914007793e0, \\ &&& 56e0d6cf231d4119001ac423, 572754cd5951b619008f8867,  570d4a6bfed7b91900d45e13, 57284b904b864d19001648e5, \\ &&& 5726cc11dd62a815002e9086, 572966ebaf94a219006aa392, 5726c3da708984140094d0d9, 57277bfc708984140094dedd, \\ &&& 572747dd5951b619008f87aa, 57107c24a58dae1900cd69ea,  571cdcb85efbb31900334e0d, 56e10e73cd28a01900c674ec, \\ &&& 5726c0c5dd62a815002e8f79, 5725f39638643c19005acefb,  5726bcde708984140094cfc2, 56e74bf937bdd419002c3e36, \\ &&& 56d997cddc89441400fdb586, 5728349dff5b5019007d9f01,  573011de04bcaa1900d770fc, 57274f49f1498d1400e8f620, \\ &&& 57376df3c3c5551400e51ed7, 5726bd655951b619008f7ca3,  5733266d4776f41900660714, 5725bc0338643c19005acc12, \\ &&& 572ff760b2c2fd1400568679, 572fbfa504bcaa1900d76c73,  5726938af1498d1400e8e448, 5728ef8d2ca10214002daac3, \\ &&& 5728f3724b864d1900165119, 56f85bb8aef2371900626011 \\
 \hline
 Oracle Error Analysis & \ref{tab:oracle-error} & SQuAD & 57376df3c3c5551400e51eda, 5726a00cf1498d1400e8e551, 5725f00938643c19005aceda, 573361404776f4190066093c, \\
&&& 571bb2269499d21900609cac, 571cebc05efbb31900334e4c, 56d7096b0d65d214001982fd, 5732b6b5328d981900602025, \\ &&& 56beb6533aeaaa14008c928e, 5729e1101d04691400779641, 56d601e41c85041400946ecf, 57115b8b50c2381900b54a8b, \\ &&& 56e74d1f00c9c71400d76f70, 5728245b2ca10214002d9ed6, 5725c2a038643c19005acc6f, 57376828c3c5551400e51eba, \\ &&& 573403394776f419006616df, 5728d7c54b864d1900164f50, 57265aaf5951b619008f706e, 5728151b4b864d1900164429, \\ &&& 57060cc352bb89140068980e, 5726e08e5951b619008f8110, 57266cc9f1498d1400e8df52, 57273455f1498d1400e8f48e, \\ &&& 572972f46aef051400154ef3, 5727482bf1498d1400e8f5a6, 57293f8a6aef051400154bde, 5726f8abf1498d1400e8f166, \\ &&& 5737a9afc3c5551400e51f63, 570614ff52bb89140068988b, 56bebd713aeaaa14008c9331, 57060a1175f01819005e78d3, \\ &&& 5737a9afc3c5551400e51f62, 57284618ff5b5019007da0a9, 570960cf200fba1400367f03, 572822233acd2414000df556, \\ &&& 5727b0892ca10214002d93ea, 57268525dd62a815002e8809, 57274b35f1498d1400e8f5d6, 56d98c53dc89441400fdb545, \\ &&& 5727ec062ca10214002d99b8, 57274e975951b619008f87fa, 572686fc708984140094c8e8, 572929d56aef051400154b0c, \\ &&& 570d30fdfed7b91900d45ce3, 5726b1d95951b619008f7ad0, 56de41504396321400ee2714, 5726472bdd62a815002e8046, \\ &&& 5727d3843acd2414000ded6b, 5726e9c65951b619008f8247 \\
 \hline
\topk~vs \dyn & \ref{tab:dyn-over-topk} & SQuAD & 56e7504437bdd419002c3e5b \\
 \hline
\full~vs \ours & \ref{tab:adv-analysis} & SQuAD-Adversarial &
56bf53e73aeaaa14008c95cc-high-conf-turk0, 56dfac8e231d4119001abc5b-high-conf-turk0
 \\
 \hline
Context Analysis & \ref{tab:triviaqa-motivation} & TriviaQA & qb 4446, wh 1933, qw 3445, qw 169, qz 2430, sfq 25261, qb 8010, qb 2880, qb 370, sfq 8018, \\
&&& sfq 4789, qz 1032, qz 603, sfq 7091, odql 10315, dpql 3949, odql 921, qb 6073, sfq 13685, bt 4547 \\
&&& sfq 23524, qw 446, jp 3302, jp 2305, tb 1951, qw 10268, bt 189, qw 14470, jp 3059, qw 12135, \\
&&& qb 7921, sfq 2723, odql 2243, qw 7457, dpql 4590,
 sfq 3509, bt 2065, qf 2092, qb 10019, sfq 14351,\\
&&& bb 4422, jp 3321, sfq 12682, sfq 13224, sfq 4027,
 qw 12518, qz 2135, qw 1983, sfq 26249, sfq 19992\\
 \hline
Error Analysis & \ref{tab:squad-analysis-more} & SQuAD & 56f84485aef2371900625f74, 56bf38383aeaaa14008c956e, 5726bb64591b619008f7c3c \\
 \hline
\end{tabular}
}
\end{center}
\reduce
\caption{QuestionIDs of samples used for human studies and analyses.} 
\label{tab:sample-index}
\end{table*}

\paragraph{Error analyses}
We compare the error cases (in exact match (EM)) of \full~and \ours. The left-most \venndiagram in Figure~\ref{fig:squad-error-analysis}  shows that \ours~is able to answer correctly to more than $97\%$ of the questions answered correctly by \full. The other two diagrams in Figure~\ref{fig:squad-error-analysis}  shows the error cases of each model, broken down by the sentence where the model's prediction is from.

Table~\ref{tab:squad-analysis-more} shows error cases on SQuAD, which \ours~fails to answer correctly. In the first two examples, our sentence selector choose the oracle sentence, but the QA model fails to answer correctly, either predicting the wrong answer from the oracle sentence, or predicting the answer from the wrong sentence. In the last example, our sentence selector fails to choose the oracle sentence. We conjecture that the selector rather chooses the sentences containing the word `the Jin dynasty', which leads to the failure in selection.
\section{Full Results on TriviaQA and SQuAD-Open}\label{sec:app-results}Table~\ref{tab:triviaqa-result} and Table~\ref{tab:squad-open-result} show full results on TriviaQA (Wikipedia) and SQuAD-Open, respectively.

\ours~obtains higher F1 and EM over \full, with the inference speedup of up to $13.8\times$.
In addition, outperforms the published state-of-the-art on both TriviaQA (Wikipedia) and SQuAD-Open, by 5.2 F1 and 4.9 EM, respectively.
\section{Samples on SQuAD, TriviaQA and SQuAD-Adversarial}\label{sec:app-samples}Table~\ref{tab:sample-index} shows the full index of samples used for human studies and analyses.

\end{document}